\documentclass[letterpaper]{article} 
\usepackage{aaai25}  
\usepackage{times}  
\usepackage{helvet}  
\usepackage{courier}  
\usepackage[hyphens]{url}  
\usepackage{graphicx} 
\usepackage{natbib}  
\usepackage{caption} 
\usepackage{algorithm}
\usepackage{algorithmic}

\usepackage{subcaption}
\usepackage{amsmath}
\usepackage{amsfonts}
\usepackage{booktabs}
\usepackage{multirow}
\usepackage{newfloat}
\usepackage{listings}
\DeclareCaptionStyle{ruled}{labelfont=normalfont,labelsep=colon,strut=off} 
\floatstyle{ruled}
\newfloat{listing}{tb}{lst}{}
\floatname{listing}{Listing}
\title{Compositional Zero-Shot Learning with Contextualized Cues and Adaptive Contrastive Training}

\author {
    Yun Li\textsuperscript{\rm 1},
    Zhe Liu\textsuperscript{\rm 2},
    Lina Yao\textsuperscript{\rm 1}
}
\affiliations {
    \textsuperscript{\rm 1}CSIRO's Data61\\
    \textsuperscript{\rm 2}UNSW\\
    y.li@csiro.au, zheliu912@gmail.com, lina.yao@data61.csiro.au
}

\usepackage{bibentry}

\begin{document}

\maketitle

\begin{abstract}
Compositional Zero-Shot Learning (CZSL) aims to recognize unseen combinations of seen attributes and objects. Current CLIP-based methods in CZSL, despite their advancements, often fail to effectively understand and link the attributes and objects due to inherent limitations in CLIP's pretraining mechanisms.  To address these shortcomings, this paper introduces a novel framework, Understanding and Linking Attributes and Objects (ULAO) in CZSL, which comprises two innovative modules. The Understanding Attributes and Objects (UAO) module improves primitive understanding by sequential primitive prediction and leveraging recognized objects as contextual hints for attribute classification. Concurrently, the Linking Attributes and Objects (LAO) module improves the attribute-object linkage understanding through a new contrastive learning strategy that incorporates tailored hard negative generation and adaptive loss adjustments. We demonstrate our model’s superiority by showcasing its state-of-the-art performance across three benchmark datasets in both Closed-World (CW) and Open-World (OW) scenarios.
\end{abstract}

\section{Introduction}
\label{sec:intro}

Compositional and zero-shot learning are crucial facets of human cognitive ability, enabling the recognition of novel concepts by combining known ones, such as identifying a pink bear from prior knowledge of pink colors and bears \cite{lake2014towards}. This capability is essential for generalizing from limited data to a broader range of unseen combinations. Compositional Zero-Shot Learning (CZSL) emulates this aspect of human intelligence \cite{li2020symmetry,naeem2021learning,karthik2022kg,li2023distilled}. In CZSL, models are trained on a finite set of attribute-object pairs (e.g., attributes like color and objects like clothing) and are challenged to identify novel combinations of these attributes and objects not seen together during training.

Traditional vision-based methods in CZSL operate at two levels: composition and primitive (i.e., attribute or object). At the composition level, methods project images and composition labels into a shared feature space to conduct similarity searches \cite{xu2021zero,naeem2021learning,wei2019adversarial}. At the primitive level, they decompose the task into separate attribute and object classification, an then recompose them to recognize compositions \cite{karthik2022kg,liu2023pami,li2023distilled}. Recently, these approaches have been significantly enhanced by integrating large pre-trained vision-language models (VLMs), such as CLIP \cite{radford2021learning}. For example, CSP \cite{csp2023} employs a structured prompt format with CLIP to recognize novel compositions in CZSL. Leveraging VLMs' strong image and text aligning capabilities, these CLIP-based methods \cite{ wang2023hierarchical,li2024context} have markedly improved performance over traditional methods.

\begin{figure}[t]
\centering
  \begin{subfigure}{0.9\linewidth}
  \centering
  \includegraphics[width=0.9\textwidth]{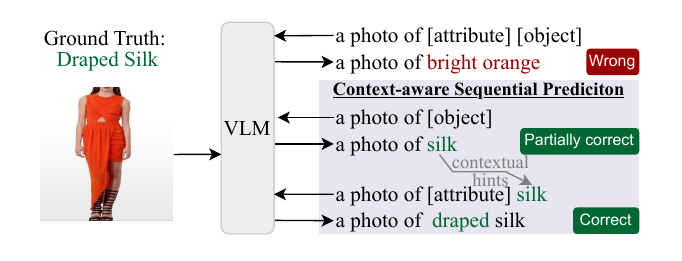}
    \caption{Misunderstanding of Attributes and Objects.}
    \label{fig:intro1}
  \end{subfigure} 
  \begin{subfigure}{0.9\linewidth}
  \centering
  \includegraphics[width=0.9\textwidth]{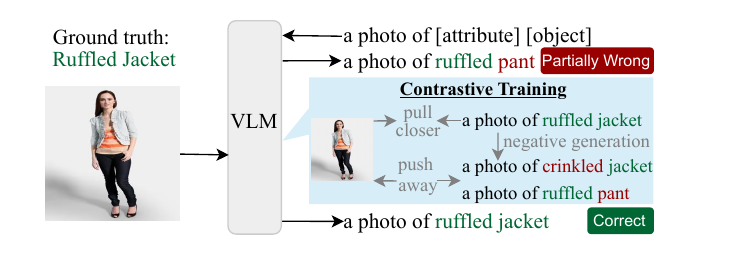}
    \caption{Incorrect Attribute-Object Linking.}
    \label{fig:intro2}
  \end{subfigure}
  \caption{Failure cases of using CLIP-based models to solve CZSL and how our model fixes them by context-aware sequential learning and contrastive learning.}
  \label{fig:intro}
\end{figure}

Despite advancements with CLIP-based methods in CZSL, they often focus on prompt-engineering or fine-tuning adaptation layers \cite{wang2023hierarchical,bao2023prompting,zheng2024caila} but overlook a fundamental flaw: CLIP's inadequate understanding of primitives and compositions \cite{yuksekgonul2022and,momeni2023verbs,hsieh2023sugarcrepe}. For instance, in Fig~\ref{fig:intro1}, an image with the label Draped-Silk is misinterpreted by the model as ``a photo of bright orange,'' where `orange,' the color describing the silk dress, is mistakenly recognized as the main object. Similarly, in Fig~\ref{fig:intro2}, a Ruffled-Jacket is incorrectly predicted as Ruffled-Pant, with the attribute `ruffled' erroneously linked to `pant' instead of `jacket.'
These examples illustrate CLIP's limitations in understanding primitives and accurately linking attributes to their objects, a problem stemming from their pretraining on cross-modal image-text retrieval tasks. These tasks rarely present composition-related challenges, allowing them to bypass in-depth learning of primitives and their relationships \cite{yuksekgonul2022and}. This limitation reduces the effectiveness of CLIP-based methods in CZSL, where precise comprehension and assembly of primitives are crucial.

In this paper, we introduce a new framework, Understanding and Linking Attributes and Objects (ULAO) in CZSL, consisting of two modules designed to resolve the problems of misunderstanding and mislinking primitives, respectively. The first Understanding Attributes and Objects (UAO) module deepens object understanding by initially only focusing on object recognition. This focus, as shown in Fig~\ref{fig:intro1}, reduces confusion from attribute-related features and thus predicts `Silk' correctly. Recognized objects then provide contextual hints for attribute classification, improving understanding of attributes with varied contextuality, for example, how attributes such as `Old' look differently when associated with different objects such as `Car' and `Cat.'

However, while UAO alleviates misunderstandings, challenges with incorrect attribute-object linkages persist. To address this, our framework introduces the Linking Attributes and Objects (LAO) module to learn primitive linkages contrastively. Traditional contrastive methods in CZSL~\cite{li2022siamese,yang2023dual} randomly select image pairs that share the same attribute or object, such as images of Ruffled-Cake and Ruffled-Jacket, pulling close their visual representations to teach the model what `Ruffled' looks like, without explicitly learning which object in the image links to Ruffled.
In contrast, our LAO employs a novel textual-negative generating strategy to enhance compositional understanding. It generates textual hard negatives by pairing the ground truth attributes or objects with those not present in the ground truth but predicted with the highest likelihood by the UAO module. For example,  as in Fig~\ref{fig:intro2}, `Pant' is the most probable object predicted by UAO. Thus, a hard negative Ruffled-Pant is generated. The contrastive loss is then trained to ensure that the similarity between the image and its ground truth (e.g., Ruffled-Jacket) exceeds the similarity between the image and any hard negative pair (e.g., Ruffled-Pant) by an adaptive threshold. This forces LAO to learn how to correctly link attributes like `Ruffled' to their objects, thus promoting accurate compositional learning.

In summary, our contributions include 1) the introduction of the ULAO framework for CZSL with UAO to enhance primitive understanding by sequential primitive prediction and leveraging recognized objects as contextual hints, 2) the development of LAO, which generates hard negatives based on prediction of UAO and use contrastive training with adaptive similarity thresholds to ensure precise attribute-object linkage and more accurate compositional learning, 3) empirical validation showing that our approach achieves superior performance than existing CLIP-based CZSL methods.

\section{Related work}

\textbf{Compositional Zero-shot Learning (CZSL).}  In CZSL, two main strategies are adopted for inferring unseen attribute-object compositions. The first strategy involves separately predicting attributes and objects, then combining these predictions to form composite labels \cite{naeem2021learning,purushwalkam2019task,li2020symmetry,liu2023pami}, with methods like Liu et al. \shortcite{liu2023pami} enhancing prediction by modeling the contextual and feasibility relationships between attributes and objects. The second strategy directly predicts compositions by mapping images and textual labels into a shared space to identify the closest composition matches \cite{xu2021relation,saini2022disentangling,li2022siamese}. For example, Mancini et al. \shortcite{mancini2022learning} utilize graph convolutional networks to propagate information across seen and unseen concepts.
Additionally, traditional methods often improve discrimination capabilities through contrastive training, typically by generating negatives randomly to refine visual and semantic representations \cite{li2022siamese,yang2023dual}. Our approach innovates on this by introducing a new strategy for generating hard negatives and implementing an adaptive loss to learn primitive linkages.

Recent works in CZSL, employing pre-trained VLMs to enhance image and text alignment \cite{lu2023decomposed,csp2023,wang2023hierarchical,Huang2024Troika}. Pioneering efforts by Nayak et al. \shortcite{csp2023} introduced CLIP’s prompting mechanism to CZSL. More recent methods deploy multi-branch models that predict attributes, objects, and compositions at the same time using VLMs \cite{wang2023hierarchical,Huang2024Troika,li2024context}. Although similar in using a multi-branch structure, our ULAO framework distinctively identifies and addresses the inherent challenges of using CLIP in CZSL, i.e., the misunderstanding and mislinking of primitives, which is neglected by previous works.

\textbf{Vision-Language Compositionality.}  
Recent research reveals that VLMs such as CLIP, although good at processing multimodal data, often fail to grasp complex compositions due to their tendency to represent information in a "bag of words" style \cite{yuksekgonul2022and,momeni2023verbs,hsieh2023sugarcrepe}. This flaw is pronounced in benchmarks assessing VLMs across various compositional dimensions, including relations, attributes, and objects. Specific benchmarks like ARO highlight attributes and relations \cite{yuksekgonul2022and}, and Winoground evaluates reasoning capabilities \cite{thrush2022winoground}. Yuksekgonul et al. \shortcite{yuksekgonul2022and} suggest enhancing negative sample generation by rearranging words in captions to combat these deficiencies. Alternatives aimed at improving caption quality involve constructing scene graphs \cite{singh2023coarse} or integrating additional caption models \cite{doveh2024dense}, yet these are designed for image-text retrieval tasks and typically require retraining or fine-tuning CLIP, unsuitable for CZSL due to limited label length and limited data volume. Our work, ULAO, is specifically crafted to tackle the compositional challenges of CLIP in CZSL settings.

\begin{figure*}[t]
\centering
\includegraphics[width=0.9\textwidth]{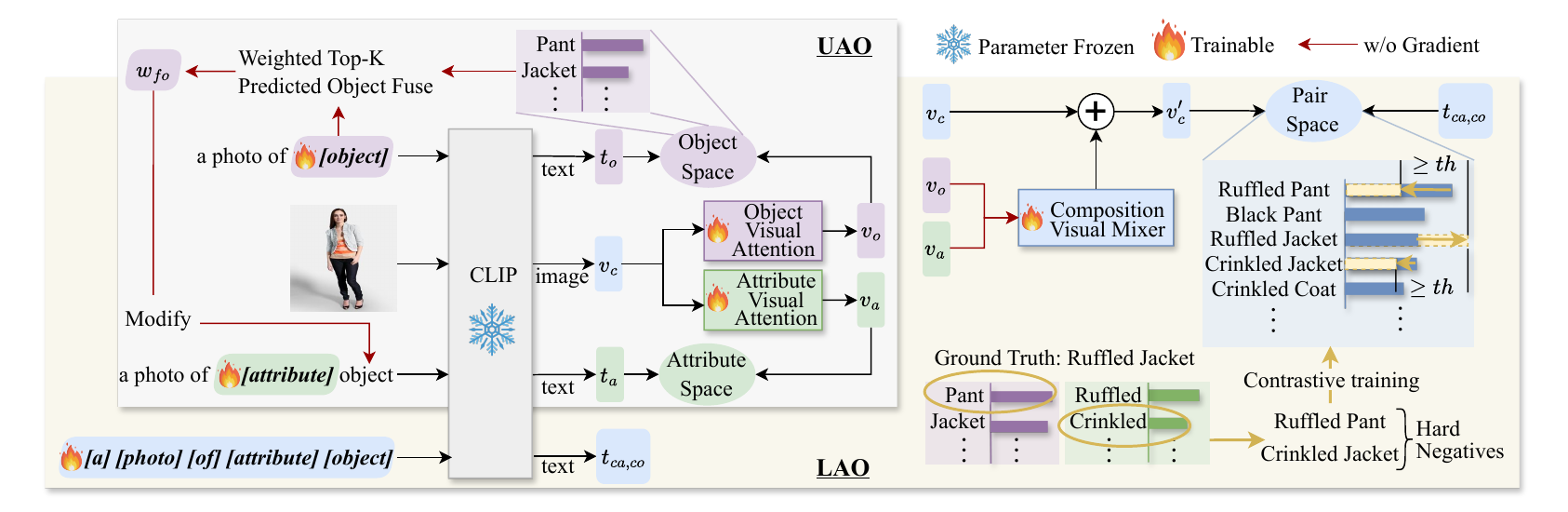}
\caption{Model Overview. The proposed ULAO equips UAO and LAO to understand and link attributes and objects.}
\label{fig:model}
\end{figure*}

\section{Method}

\textbf{Task formulation and definitions.} In CZSL, images are categorized as attribute-object pairs, where each attribute $a \in \mathcal{A}$ and object $o \in \mathcal{O}$ contribute to the composition label space $Y = \mathcal{A} \times \mathcal{O}$. This label space encompasses all combinations of attributes and objects, some of which may be infeasible in real-world scenarios, such as Tasty-Wood.
Training in CZSL targets only a subset of this label space, known as seen compositions $Y^{\mathcal{S}} = \{(x, y) | x \in X^{\mathcal{S}}, y \in Y^{\mathcal{S}}\}$, where each image $x$ is associated with single attribute-object pair $y = (a, o)$. During testing, the model needs to classify images from both these seen compositions and unseen compositions $Y^{\mathcal{U}}$, which do not overlap with $Y^{\mathcal{S}}$ and are not present during training. The testing scenarios are distinct: in Closed-World (CW), the label space is restricted to $y \in Y^{\mathcal{S}} \cup Y^{\mathcal{U}}$, with unseen compositions $Y^{\mathcal{U}}$ provided before testing. In contrast, in Open-World (OW), no specific unseen compositions are informed beforehand; thus, the model needs to predict in the entire label space $Y$, covering all attribute-object combinations, feasible or not.

\textbf{Model overview.} Our ULAO framework addresses the limitations of existing CLIP-based methods by introducing two innovative modules: the Understanding Attributes and Objects (UAO) module and the Linking Attributes and Objects (LAO), illustrated in Fig~\ref{fig:model}. The UAO module enhances primitive classification through a sequential prediction strategy. This strategy minimizes interference from overlapping attribute features for object predictions and leverages recognized objects as contextual cues for subsequent attribute predictions, thus refining the understanding of attributes.
The LAO module addresses linkage errors between attributes and objects, an issue inadequately resolved by current methodologies. It employs a contrastive learning strategy that is distinct from traditional approaches, utilizing dynamically generated hard negatives based on real-time prediction probabilities. This strategy ensures robust learning of accurate attribute-object associations.

\subsection{Understanding attributes and objects (UAO)}

Following literatures~\cite{wang2023hierarchical}, we utilize CLIP's visual and textual encoders to map images and textual labels into a unified representational space. Here, we compute the similarity to determine the most compatible label for each image, accommodating both complete pair labels and individual attribute or object labels for different prediction tasks. We fix the visual and textual encoders without fine-tuning, in line with findings from Huang et al. \shortcite{Huang2024Troika} that fine-tuning CLIP in CZSL can degrade performance. 

With the encoders fixed, to tailor the visual outputs for our specific tasks, after obtaining the visual representation $v_{c}$ from the visual encoder $f_{v}$ of CLIP, we apply two distinct visual attention modules, $f_a$ and $f_o$, to derive attribute-specific and object-specific visual representations respectively:
\begin{equation}
    v_{c} = f_{v}(x), \qquad v_{a}=f_{a}(v_{c}), \qquad  v_{o}=f_{o}(v_{c})
\end{equation}

For the textual side, we employ soft prompts that are learnable during training to produce textual representations that fit the CZSL task. Specifically, for object classification, the prompt is structured as \underline{\textit{a photo of [\textbf{object}]}}. We denote the prompts as $\theta_{o} = \{e_{0}, e_{1}, e_{2}, w_{o}\}$, where $w_{o}$, initialized with the word embeddings of object labels, is updated during training. The remaining elements $e_{0}, e_{1}, e_{2}$ represent the prefix ``a photo of'' and are fixed during training. This prompt $\theta_{o}$ is then processed by the textual encoder $f_{t}$ of CLIP to generate the textual representations: $t_{o} = f_{t}(\theta_{o})$.
Finally, we normalize the representations $t_{o}=\frac{t_{o}}{\| t_{o} \|}$, and $v_{o}=\frac{v_{o}}{\| v_{o} \|}$, and compute object probabilities as follows:
\begin{equation}
    p(o|x)=\frac{exp(v_{o}\cdot t_{o}/ \tau)}{\sum_{\hat{o}\in \mathcal{O}}exp(v_{o}\cdot t_{\hat{o}}/ \tau)}
\end{equation}
where $\tau$ is a temperature parameter from CLIP.

After computing object probabilities, we refine the attribute classification prompts. Existing works \cite{wang2023hierarchical, li2024context} typically format attribute prompts as \underline{\textit{a photo of [\textbf{attribute}] object}} ($\theta_{a} = \{e_{0}, e_{1}, e_{2}, w_{a}, e_{3}\}$). However, to incorporate contextuality into attribute perception, we dynamically adjust the fixed word \textit{object} ($ e_{3}$) in $\theta_{a}$ for each image $x$ to include $x$'s most probable object labels as contextual cues.
To address uncertainties in object predictions, we incorporate the top-$k$ predicted objects and compute their collective influence, $w_{fo}$, as a weighted sum of their embeddings to modify the fixed \textit{object} ($ e_{3}$) to get $e_3'$:
\begin{equation}
    w_{fo} = \sum_{i=1}^{k} p(o_i|x) \cdot w_{o_i}, \qquad e'_{3} = (1-r_{k})e_{3} + r_{k} \cdot w_{fo}
\end{equation}
where the weight in $w_{fo}$ is set as $p(o_i|x)$ (the probability of object $o_i$), and $w_{o_i}$ is embedding of $o_i$. The hyperparameters $k$ and $r_k$ control the extent of this contextual integration, i.e., how many object predictions are incorporated and the incorporation ratio. Then, we use $e_3'$ to replace $e_3$ to dynamically form attribute prompts. Note that we detach gradients from $w_{fo}$ during training, preventing attribute classification from influencing object recognition and thus avoiding cases like confusing `Silk' with `Orange' in Fig~\ref{fig:intro1}.

Similarly to computing object probabilities, we compute the attribute probabilities after normalization:
\begin{equation}
p(a|x)=\frac{exp(v_{a}\cdot t_{a}/ \tau)}{\sum_{\hat{a}\in \mathcal{A}}exp(v_{a}\cdot t_{\hat{a}}/ \tau)}
\end{equation}

We then calculate the probability of attribute-object compositions as the product of attribute and object probabilities, $p(y=(a,o)|x) = p(a|x)p(o|x)$. The final loss function for the UAO module integrates all three probabilities as follows, where $r_{ao}\in \mathbb{R}$ is the loss coefficient:
\begin{equation}
\begin{aligned}
\mathcal{L}_{\mathit{UAO}} = -\frac{1}{|\mathcal{S}|} \sum_{(x,y) \in \mathcal{S}} &\left( \log p(a|x)  + \log p(o|x) \right. \\
&\left.+ r_{ao} \log p((a,o)|x) \right)
\end{aligned}
\end{equation}

\subsection{Linking attributes and objects (LAO)}
\label{sec:lao}

Building upon the UAO module's enhanced primitive understanding, the LAO module aims to refine the  primitive linkages. For example, consider an image from Fig~\ref{fig:model} that depicts a female wearing a ruffled jacket and black pants. The correct label, Ruffled-Jacket, gets mispredicted as Ruffled-Pant due to linkage errors in direct composition prediction using CLIP.
To address such errors, LAO generates hard negatives by substituting either the attribute or object of the true label with the most probable mispredictions identified by the UAO module. In this instance, the hard negatives generated are Crinkled-Jacket and Ruffled-Pant. During training, LAO ensures that the similarity between the image and its true label (Ruffled-Jacket) surpasses the similarity between the image and any of the hard negative labels by a dynamically adapted threshold. This strategy not only increases the probability of the correct label but also decreases the likelihood of similar but incorrect labels (Crinkled-Jacket) and corrects the attribute-object linkage by penalizing incorrect pairings like Ruffled-Pant.  Through adaptive threshold adjustments, LAO enhances the model's accuracy in predicting compositions, differentiating between similar primitives, and confirming proper attribute-to-object connections. Next, we introduce the detailed implementation.

\textbf{Similarity calculation.} The LAO module introduces a composition prediction branch to obtain similarities between image and composition labels directly. For this branch, we integrate visual features $v_c$, $v_a$, and $v_o$ obtained in UAO through a Composition Visual Mixer $f_m$. This integration forms a combined visual representation: $v'_c = v_c + r_{m} \cdot f_{m}(v_a, v_o)$, where $r_m$ is a weighting parameter.
For textual representation, following \cite{lu2023decomposed}, we utilize a fully learnable prompt $\theta_{c} = \{w_{0}, w_{1}, w_{2}, w_{ca}, w_{co}\}$, formatted as \underline{\textit{[a] [photo] [of] [\textbf{attribute}] [\textbf{object}]}}. The prompt is processed through CLIP’s textual encoder to obtain $t_{ca, co} = f_{t}(\theta_{c})$. Similar to the UAO module, we normalize both visual and textual representations in LAO: $t_{ca, co} = \frac{t_{ca, co}}{\left\| t_{ca, co} \right\|}$ and $v'_{c} = \frac{v'_{c}}{\left\| v'_{c} \right\|}$. The (image, composition-label) similarity is then calculated as $s(x, (ca,co)) = v'_{c} \cdot t_{ca, co}/\tau$ and the composition prediction is as follows:
\begin{equation}
    p(y=(ca, co)|x)=\frac{\exp(v'_{c} \cdot t_{ca, co}/ \tau)}{\sum_{(\hat{ca},\hat{co}) \in Y^{\mathcal{S}}} \exp(v'_{c} \cdot t_{\hat{ca}, \hat{co}}/ \tau)}
\end{equation}

\begin{table}[t]
\small
\centering
 \setlength{\tabcolsep}{1.5pt}
\begin{tabular}{lcc|cc|ccc}
\toprule
           &     &        & \multicolumn{2}{c|}{Training} & \multicolumn{3}{c}{Testing}                \\
Dataset    & a   & o  &seen     & image           & seen  & unseen  & image    \\
\midrule
MIT-States & 115 & 245   & 1262          & 30k          & 400  & 400(27k)    & 13k                  \\
UT-Zappos  & 16  & 12      & 83            & 23k          & 18   & 18(109)     & 3k                 \\
C-GQA      & 413 & 674  & 5592          & 27k          & 888 & 923(272k)   & 5k     \\
\bottomrule
\end{tabular}
\caption{Statistics of datasets. Numbers in brackets are the unseen search space in OW.  }
\label{tab:datasets}
\end{table}

\begin{table*}[t]
\small
\centering
\begin{tabular}{c|l|cccc|cccc|cccc}
\toprule
& & \multicolumn{4}{c|}{MIT-States}                  & \multicolumn{4}{c|}{UT-Zappos}                 & \multicolumn{4}{c}{C-GQA}\\
& & ~S~& ~U~& HM& AUC & ~S~ & ~U~& HM & AUC& ~S~& ~U~& HM&AUC\\ \midrule
\multicolumn{14}{c}{Closed-World (CW)}  \\ \midrule
\multirow{3}*{\textbf{w/o}}
&Co-CGE~\cite{mancini2022learning} & 32.1 & 28.3 & 20.0 & 6.6 & 62.3 & 66.3 & 48.1 & 33.9 & 33.3 & 14.9 & 14.4 & 4.1 \\
&SCEN~\cite{li2022siamese} & 29.9 & 25.2 & 18.4 & 5.3 & 63.5 & 63.1 & 47.8 & 32.0 & 28.9 & 25.4 & 17.5 & 5.5 \\
&CANet~\cite{wang2023learning} & 29.0 & 26.2 & 17.9 & 5.4 & 61.0 & 66.3 & 47.3 & 33.1 & 30.0 & 13.2 & 14.5 & 3.3 \\
\midrule
\multirow{10}*{\textbf{w}}
& CLIP~\cite{radford2021learning} & 30.2 & 46.0 & 26.1 & 11.0 & 15.8 & 49.1 & 15.6 & 5.0  & 7.5  & 25.0 & 8.6  & 1.4  \\
&Co-CGE~\cite{mancini2022learning} & 46.7 & 45.9 & 33.1 & 17.0 & 63.4 & 71.3 & 49.7 & 36.3 & 34.1 & 21.2 & 18.9 & 5.7  \\
&CoOP~\cite{zhou2022learning} & 34.4 & 47.6 & 29.8 & 13.5 & 52.1 & 49.3 & 34.6 & 18.8 & 20.5 & 26.8 & 17.1 & 4.4 \\
&CSP~\cite{csp2023} & 46.6 & 49.9 & 36.3 & 19.4 & 64.2 & 66.2 & 46.6 & 33.0 & 28.8 & 26.8 & 20.5 & 6.2 \\
&HPL~\cite{wang2023hierarchical} & 47.5 & 50.6 & 37.3 & 20.2 & 63.0 & 68.8 & 48.2 & 35.0 & 30.8 & 28.4 & 22.4 & 7.2 \\
&DFSP~\cite{lu2023decomposed} & 46.9 & 52.0 & 37.3 & 20.6 & \underline{66.7} & 71.7 & 47.2 & 36.0 & 38.2 & 32.0 & 27.1 & 10.5 \\
&GIPCOL~(Xu et al.~\citeyear{xu2024gipcol}) & 48.5 & 49.6 & 36.6 & 19.9 & 65.0 & 68.5 & 48.8 & 36.2 & 31.9 & 28.4 & 22.5 & 7.1 \\
&ProLT~\cite{jiang2024revealing} & 49.1 & 51.0 & 38.2 & 21.1 & 66.0 & 70.1 & 49.4 & 36.1 & \textbf{39.5} & 32.9 & 27.7 & 11.0 \\
&CDSCZSL~\cite{li2024context} & \underline{50.3} & \textbf{52.9} & \underline{39.2 }& \underline{22.4} & 63.9 & \underline{74.8} & \underline{52.7} & \underline{39.5} & 38.3 & \underline{34.2} & \underline{28.1} & \underline{11.1} \\
\cmidrule(lr){2-14}
&ULAO (ours) & \textbf{51.4} & \underline{52.3} & \textbf{39.3} & \textbf{22.7} & \textbf{68.1} & \textbf{75.2} & \textbf{57.2} & \textbf{44.3} & \underline{39.2} & \textbf{34.5} & \textbf{28.7} & \textbf{11.7} \\
\midrule
\multicolumn{14}{c}{Open-World (OW)}  \\ \midrule
\multirow{3}*{\textbf{w/o}} &Co-CGE~\cite{mancini2022learning} & 30.3 & 11.2 & 10.7 & 2.3 & 61.2 & 45.8 & 40.8 & 23.3 & 32.1 & 3.0 & 4.8 & 0.78 \\
&SAD-SP~\cite{liu2023pami} & 29.1 & 7.6 & 7.8 & 1.4 & 63.1 & 54.7 & 44.0 & 28.4 & 31.0 & 3.9 & 5.9 & 1.00 \\
&DRANet~\cite{li2023distilled} & 29.8 & 7.8 & 7.9 & 1.5 & 65.1 & 54.3 & 44.0 & 28.8 & 31.3 & 3.9 & 6.0 & 1.05 \\
\midrule
\multirow{10}*{\textbf{w}}
&CLIP~\cite{radford2021learning}& 30.1 & 14.3 & 12.8 & 3.0  & 15.7 & 20.6 & 11.2 & 2.2  & 7.5  & 4.6  & 4.0  & 0.27  \\
&Co-CGE~\cite{mancini2022learning}& 38.1 & 20.0 & 17.7 & 5.6  & 59.9 & 56.2 & 45.3 & 28.4 & 33.2 & 3.9  & 5.3  & 0.91\\
&CoOP~\cite{zhou2022learning} & 34.6 & 9.3 & 12.3 & 2.8 & 52.1 & 31.5 & 28.9 & 13.2 & 21.0 & 4.6 & 5.5 & 0.70 \\
&CSP~\cite{csp2023} & 46.3 & 15.7 & 17.4 & 5.7 & 64.1 & 44.1 & 38.9 & 22.7 & 28.7 & 5.2 & 6.9 & 1.20 \\
&HPL~\cite{wang2023hierarchical} & 46.4 & 18.9 & 19.8 & 6.9 & 63.4 & 48.1 & 40.2 & 24.6 & 30.1 & 5.8 & 7.5 & 1.37 \\
&DFSP~\cite{lu2023decomposed} & 47.5 & 18.5 & 19.3 & 6.8 & \underline{66.8} & 60.0 & 44.0 & 30.3 & 38.3 &  7.2 & 10.4 & 2.40 \\
&GPICOL~(Xu et al.~\citeyear{xu2024gipcol}) & 48.5 & 16.0 & 17.9 & 6.3 & 65.0 & 45.0 & 40.1 & 23.5 & 31.6 & 5.5 & 7.3 & 1.30 \\
&Troika~\cite{Huang2024Troika} & 48.8 & 18.7 & 20.1 & 7.2 & 66.4 & 61.2 & 47.8 & \underline{33.0} & \textbf{40.8} & 7.9 & 10.9 & \underline{2.70} \\
&CDSCZSL~\cite{li2024context} & \underline{49.4} & \textbf{21.8} & \underline{22.1} & \underline{8.5} & 64.7 & \underline{61.3} & \underline{48.2} & 32.3 & 37.6 & \underline{8.2} & \underline{11.6} & 2.68 \\
\cmidrule(lr){2-14}
&ULAO (ours) & \textbf{51.3} & \underline{21.6} &\textbf{ 22.4} & \textbf{8.8} & \textbf{68.1} & \textbf{62.8} & \textbf{50.0} & \textbf{35.0} & \underline{39.3} & \textbf{9.4} & \textbf{12.8} & \textbf{3.23} \\
\bottomrule
\end{tabular}
\caption{Model performance in CW and OW. We use \textbf{w} and \textbf{w/o} to indicate whether models adopt CLIP as visual and language encoders. The best and second best results are in \textbf{bold} and \underline{underlined}.}
\label{tab:main}
\end{table*}

\textbf{Hard negative generation.}
As previously outlined, hard negatives are constructed using the probabilities of attributes and objects calculated by UAO. For an image $x$ with ground truth label $(ca, co)$, we generate its negative set $\mathcal{N}= \{(a', co), (ca, o')\}$ as follows:
\begin{align}
a' &= \underset{a' \in \mathcal{A} \setminus \{a\}, (a',co) \in Y^{\mathcal{S}}}{\arg\max} \, p(a'|x), \\
o' &= \underset{o' \in \mathcal{O} \setminus \{o\}, (ca,o') \in Y^{\mathcal{S}}}{\arg\max} \, p(o'|x).
\end{align}
Note that the generated negatives should belong to the seen composition set $\mathcal{S}$ to ensure we can obtain their similarity with images from the composition prediction branch.

\textbf{Contrastive Loss with Adaptive Threshold.} To enhance the model's ability to discern between correct image-label pairs and hard negatives, we employ a contrastive loss function that optimizes the difference in similarity to exceed an adaptive threshold $\mathit{th}$:
\begin{align}
\mathcal{L}_{\text{cntr}} &= \sum_{(x, (ca, co)) \in B} \sum_{(ca', co') \in \mathcal{N}} \max\Big(0,  \notag \\
&\quad s(x, (ca', co')) - s(x, (ca, co)) + \mathit{th}_{ca', co'}\Big)
\end{align}

Inspired by recent advances in curriculum learning methods (Zhang et al.~\citeyear{zhang2024contrasting}), we initialize the thresholds at zero and adapt them based on the average observed differences in similarities during training. Specifically, the adaptive threshold for negatives at time step $t$ is updated as:
\small{
\begin{equation}
\begin{split}
\Delta^{t-1}_{ca', co'} &= \frac{1}{|B|} \sum_{x\in B} \left(s^{t-1}(x, (ca, co)) - s^{t-1}(x, (ca', co'))\right)
\\
\mathit{th}^{t}_{ca', co'} &= \max \left(\mathit{th}^{t-1}_{ca', co'}, \Delta^{t-1}_{ca', co'}\right)
\end{split}
\end{equation}}

This progressive adjustment of thresholds helps the model gradually focus on negative distinctions as its discrimination ability improves. To prevent thresholds from becoming prohibitively high, which could impede learning by overly penalizing the model, we impose predefined upper bounds $up$ for $\mathit{th}$:
\begin{equation}
\mathit{th}^{t}_{ca', co'} = \min(\mathit{th}^{t}_{ca', co'}, up_{ca',co'})
\end{equation}

The overall loss for the LAO module, therefore, combines the composition prediction loss with the adaptive contrastive loss with a weighting factor $r_c$ to control the influence of the contrastive loss:
\begin{equation}
\mathcal{L}_{\mathit{LAO}} = -\frac{1}{|\mathcal{S}|} \sum_{(x, (ca, co)) \in \mathcal{S}} \log p((ca, co)|x) + r_c \mathcal{L}_{\text{cntr}}
\end{equation}

\textbf{Final prediction and training}.
We fuse predictions of the UAO and LAO modu
les to get the final prediction: $p(y|x)=\alpha p(y=(ca,co)|x)+(1-\alpha) p(a|x)p(o|x)$, where $\alpha$ is the fusing weight. The loss for the final prediction and the overall training loss for the complete ULAO model are:
\begin{align}
\mathcal{L}_{\mathit{combined}} &= -\frac{1}{|\mathcal{S}|} \sum_{(x, y) \in \mathcal{S}} \log p(y|x), \\
\mathcal{L}_{\mathit{ULAO}} &= \mathcal{L}_{\mathit{UAO}} + \mathcal{L}_{\mathit{LAO}} + \mathcal{L}_{\mathit{combined}}.
\end{align}

\section{Experiment}

\subsection{Experiment Settings}\label{sec:setting}

\textbf{Datasets and Evaluation Metrics.} 
Our evaluation spans three benchmark datasets, each presenting unique challenges: MIT-States~\cite{isola2015discovering} features diverse natural objects and fine-grained attributes; UT-Zappos~\cite{yu2014fine,yu2017semantic} focuses on a narrow range of footwear, with the primitives being highly similar; and C-GQA~\cite{naeem2021learning}, which offers a broad scope with 413 attributes and 674 objects, posing significant challenges in OW scenarios. We adopt standard splits from prior works to split the seen and unseen compositions~\cite{purushwalkam2019task,naeem2021learning}, detailed in Tab. \ref{tab:datasets}. We use best-seen accuracy (S), best-unseen accuracy (U), harmonic mean (HM), and Area Under the Curve (AUC) as performance evaluation metrics following~\cite{mancini2021open}. Among them, HM and AUC are the primary measures as they evaluate the model more comprehensively.

\textbf{Implementation Details.}
Following established practices~\cite{li2024context}, we employ CLIP~\cite{radford2021learning} for image and text encoding. Our ULAO implements the Object and Attribute Visual Attention as a single-layer Multi-head Attention and the Composition Visual Mixer as a four-layer Fully Connected Network (FCN). Optimization is performed end-to-end using the Adam optimizer~\cite{kingma2014adam} on NVIDIA A100 GPU. The provided Supplementary materials include codes, environment requirements, and detailed config files that provide hyperparameter settings, including learning rates, loss ratios, etc. 

\subsection{Comparisons with SOTAs}

Our model, ULAO, is tested against contemporary CZSL methods in both Closed-World (CW) and Open-World (OW) settings, ensuring fair comparison by utilizing consistent data splits and evaluation metrics. Most of these methods, including ours, are based on the CLIP framework with the ViT-L/14 architecture, and all employ frozen CLIP parameters during training to emphasize improvements from modeling techniques over architectural changes. 

The efficacy of ULAO is shown in Table~\ref{tab:main}, where it consistently outperforms other methods on HM and AUC across all datasets. Notably, on UT-Zappos in the CW setting, ULAO achieves AUC and HM improvements of 4.8\% and 4.5\%, respectively, and similar enhancements in OW. These gains are attributed to the LAO component’s effective differentiation of closely related attributes and objects. In the most challenging C-GQA dataset where OW search space explodes, the model excels particularly in OW scenarios, demonstrating a 19.6\% relative improvement, likely due to 1) LAO shrinking search space by excluding pairs with wrong attribute-object linkages and 2) UAO refines attribute search space by taking object information as contextual hints.

ULAO stands out even when compared to other advanced CLIP-based methods such as HPL~\cite{wang2023hierarchical}, CDSCZSL~\cite{li2024context}, and Troika~\cite{Huang2024Troika}. These methods similarly utilize multi-branch strategies for handling attributes, objects, and compositions. ULAO's superior performance over them across various settings confirms the effectiveness of our approach in enhancing the understanding and linkage of attributes and objects, proving that our gains are derived from methodological advancements rather than the underlying neural architecture.

\begin{table}[t]
    \centering
    \setlength{\tabcolsep}{1.5pt} 
        \begin{tabular}{l|cccc|cccc}
            \toprule
            & \multicolumn{4}{c|}{MIT-States} & \multicolumn{4}{c}{UT-Zappos} \\
            & ~S~ & ~U~ & HM & AUC & ~S~ & ~U~ & HM & AUC \\ 
                        \midrule
\multicolumn{9}{c}{CW}  \\ \midrule
             C & 44.1 & 51.7 & 35.8 & 19.1 & 64.3 & 66.3 & 47.3 & 33.8 \\
             C+AO & 47.1 & \textbf{53.2} & 37.5 & 21.1 & 64.5 & \underline{71.3} & 49.0 & 36.2 \\
             C+UAO & \underline{49.9} & \underline{52.8} & \underline{38.8} & \underline{22.2} & \underline{66.2} & 71.0 & \underline{52.0} & \underline{38.1} \\
             ULAO & \textbf{51.4} & 52.3 &\textbf{39.3} & \textbf{22.7} & \textbf{68.1} & \textbf{75.2} & \textbf{57.2} & \textbf{44.3} \\ 
                        \midrule
\multicolumn{9}{c}{OW}  \\ \midrule
             & & & & & & & &\\
             C & 45.8 & 16.7 & 18.3 & 6.1 & 63.2 & 52.0 & 44.1 & 27.1 \\
             C+AO & 45.0 & 21.1 & 20.6 & 7.4 & 61.9 & \underline{60.0} & 45.2 & 29.9 \\
             C+UAO & \underline{49.9} & \underline{21.5} & \underline{21.9} & \underline{8.5} & \underline{66.2} & 58.0 & \underline{46.5} & \underline{31.1} \\
             ULAO & \textbf{51.3} & \textbf{21.6} & \textbf{22.4} & \textbf{8.8} & \textbf{68.1} & \textbf{62.8} & \textbf{50.0} & \textbf{35.0} \\ 
            \bottomrule
            \end{tabular}
    \caption{Module ablation study.} 
        \label{tab:ablation}
\end{table}

\begin{figure}[t]
    \centering
    \begin{subfigure}{\linewidth}
            \centering
                \includegraphics[width=0.7\textwidth]{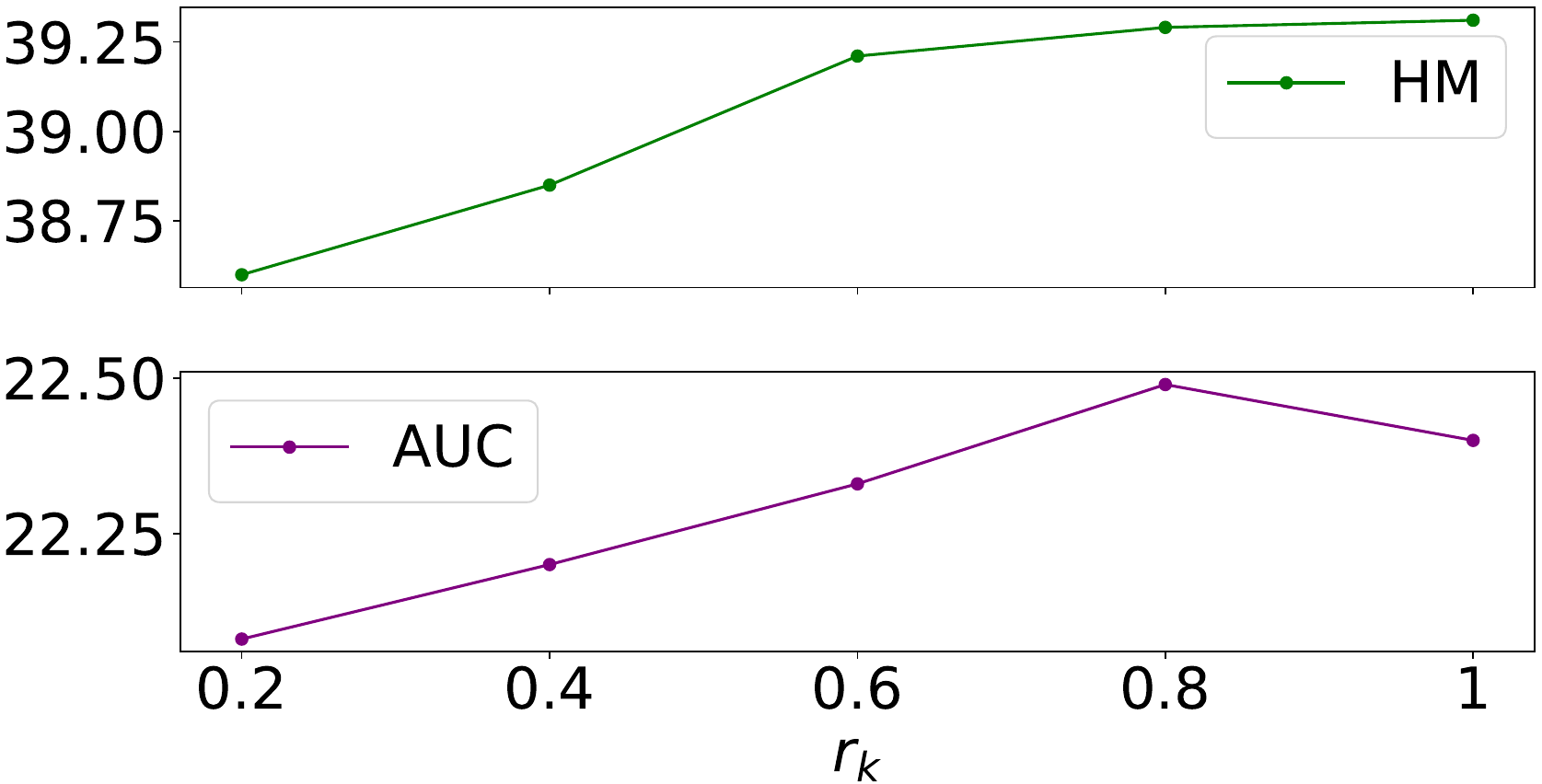}
                \caption{Study on contextual integration ratio $r_{k}$.}
                \label{fig:r_k}
    \end{subfigure}
    \begin{subfigure}{\linewidth}
            \centering
            \includegraphics[width=0.7\textwidth]{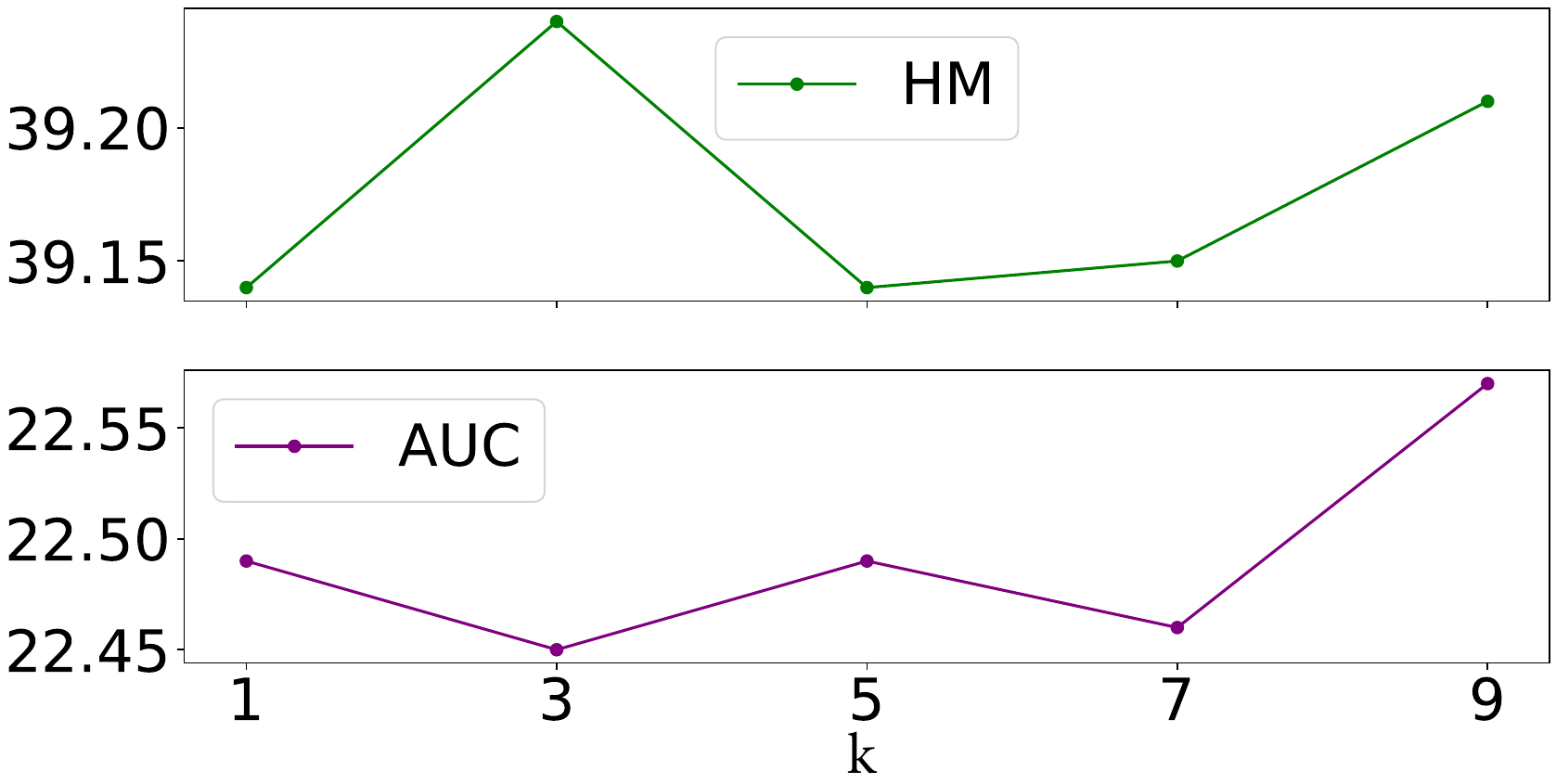}
            \caption{Study on object integration number $k$.}
            \label{fig:k}
    \end{subfigure} 
    \caption{Parameter study on UAO.} 
     \label{tab:ota}
\end{figure}

\begin{table}[t]
    \centering
    \setlength{\tabcolsep}{1.5pt}
        
        \begin{tabular}{c|c|c|c|c|c|c|cccc}
        \toprule
         \multicolumn{7}{c|}{Methods} & \multicolumn{4}{c}{MIT-States} \\
        \midrule
        C & Random & UAO & Mean & Max & Fixed & $up$ & ~S~ & ~U~ & HM & AUC \\
        \midrule
        $\checkmark$ & & & $\checkmark$ & & & & \underline{51.2} & 51.7 & 38.7 & 22.1 \\
         & $\checkmark$ & & $\checkmark$ & & & & 50.7 & \textbf{52.6} & \underline{39.0} & 22.3 \\
        &  & $\checkmark$ & $\checkmark$  & & & & 50.8 & \underline{52.3} & \textbf{39.3} & \underline{22.5} \\
         &  & $\checkmark$ &  & $\checkmark$ & & & 49.5 & \underline{52.3} & 38.7 & 21.7 \\
         &  & $\checkmark$ &  & & $\checkmark$ & & 49.8 & \underline{52.3} & 38.9 & 22.0 \\
        & & $\checkmark$ & $\checkmark$ &  & & $\checkmark$ & \textbf{51.4} & \underline{52.3} & \textbf{39.3} & \textbf{22.7} \\
        \bottomrule
        \end{tabular}
    \caption{Ablation on contrastive design on MIT-States.}
    \label{tab:contra}
\end{table}

\begin{figure}[t]
    \centering
    \includegraphics[width=0.7\linewidth]{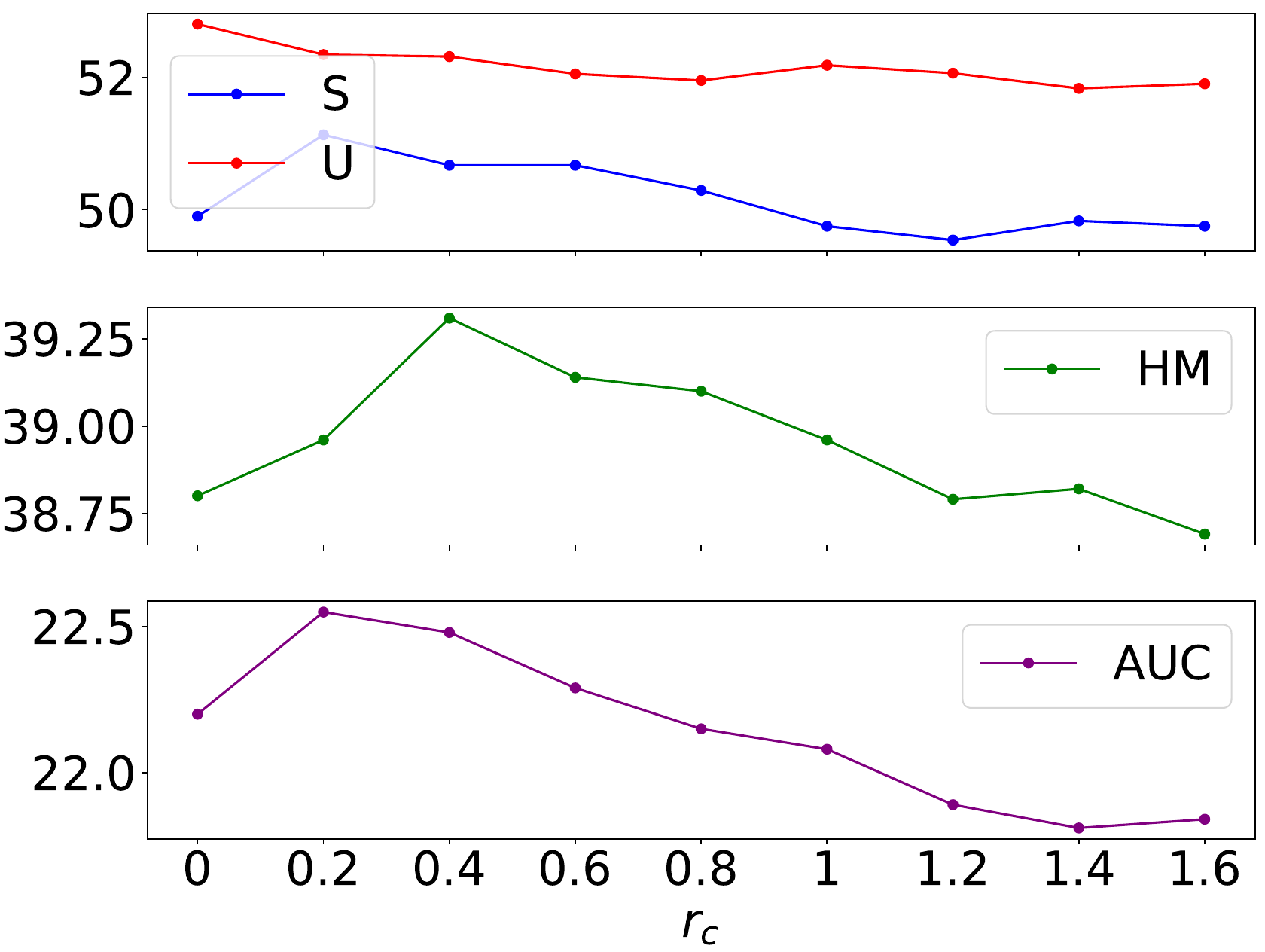}
    \caption{Study on contrastive loss ratio $r_{c}$.}
            \label{fig:r_c}
\end{figure}

\begin{figure*}[t]
    \centering
    \includegraphics[width=\textwidth]{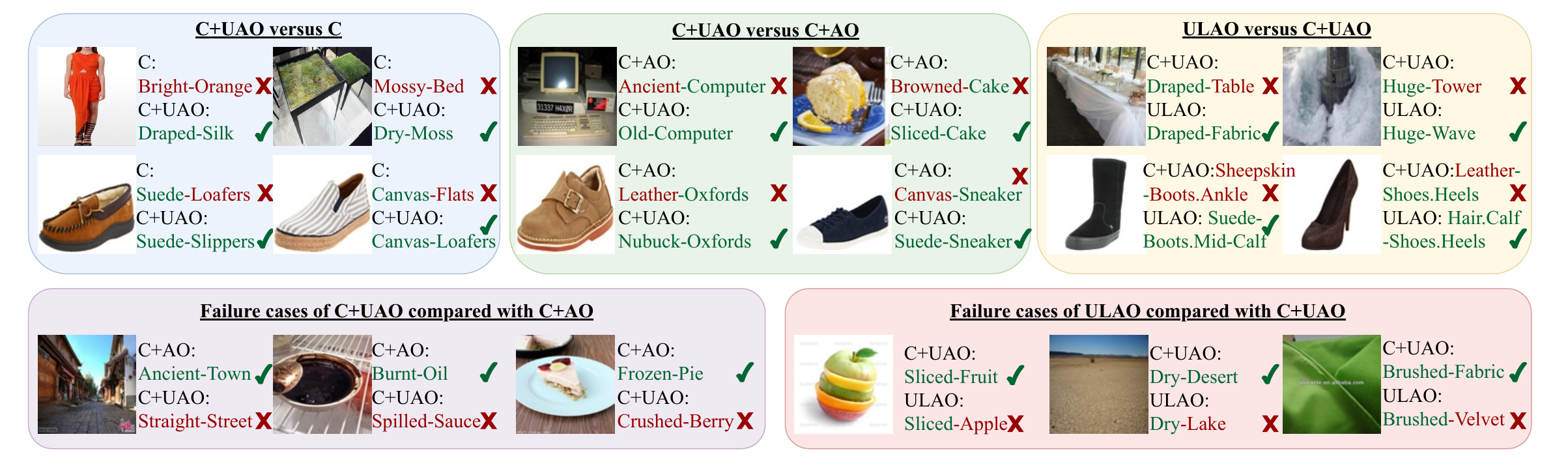}
    \caption{Qualitative results on different variants of ULAO.}
    \label{fig:qualitative}
\end{figure*}

\subsection{Ablation Study}\label{sec ablation}

\textbf{Overall module ablation}. We performed an ablation study to evaluate the contributions of each component within the ULAO framework. The variants studied are: \textit{C} (only the composition prediction branch of LAO), \textit{C+AO} (adds UAO's attribute and object prediction branches to \textit{C} without using object hints for attribute prediction), and \textit{C+UAO} (incorporates UAO's full functionality including top-k object hints but excludes LAO's contrastive loss).

Table~\ref{tab:ablation} presents the results. From \textit{C} to \textit{C+AO}, we observe enhancements in performance metrics, affirming that separate attribute and object predictions contribute significantly to primitive understanding. The improvement from \textit{C+AO} to \textit{C+UAO} demonstrates the effectiveness of integrating top-k predicted objects as contextual hints, especially evident in the HM and AUC improvements. Finally, the transition to ULAO with the contrastive loss shows further gains, underscoring the impact of refined attribute-object linkages on overall model performance. 

Our results also indicate distinct contributions of model components depending on dataset characteristics. For UT-Zappos, characterized by attributes and objects with clear yet closely related definitions (e.g., suede vs. sheepskin), the introduction of contrastive loss yields significant performance improvements. This is due to the loss's effectiveness in enhancing the model's ability to discriminate between similar categories by effectively separating confusing negatives. Conversely, the MIT-States dataset, which features more abstract and overlapping categories (e.g., tiny vs. small, beach vs. shore, or fruit vs. apple), benefits more from enhanced attribute and object understanding provided by the AO and the contextuality provided by object hints in UAO.

\textbf{Parameter study on UAO.} We analyze how the ratio of object hints $r_k$ and the number of $k$ top-$k$ predicted objects used as hints impact the performance of the UAO module. For this experiment, the seed was fixed at 0, and all other parameters were maintained at their default values to minimize randomness and ensure consistency across tests. As illustrated in Figure~\ref{fig:r_k}, increasing $r_k$ generally enhances model performance, confirming the advantageous effect of integrating more object hints. Conversely, Figure~\ref{fig:k} reveals that adjusting the number of object hints $k$ introduces some performance fluctuations, suggesting a delicate balance in the optimal number of hints used.

\textbf{Design of the Contrastive Training Strategy.} We evaluate the contrastive training strategy from three aspects in Table~\ref{tab:contra}: generating negatives, updating thresholds ($th$), and setting upper bounds ($up$). For negative generation, our approach, which selects the most probable mispredictions by UAO for substitution, was compared against using the highest probability mispredictions from the composition branch (C) and random attribute or object substitutions. Our method outperformed both alternatives, demonstrating its effectiveness.
In threshold updating, employing the mean similarity difference to update $th$ is superior to using the maximum difference or a fixed threshold. Finally, implementing an upper limit on $th$ further enhanced model performance, preventing overly punitive adjustments that could hinder effective learning. 

\textbf{Contrastive loss ratio.} We also vary the contrastive loss ratio $r_c$ and report the performance in Fig.~\ref{fig:r_c}. 
The performance increases first as $r_c$ increases, peaking around a ratio of 0.2 to 0.4. Beyond this optimal range, performance begins to decline. This trend suggests that while a moderate increase in the contrastive loss ratio can enhance discriminative learning by effectively penalizing less accurate predictions, excessive emphasis on contrastive loss may adversely affect the model's generalization capabilities on unseen compositions.


\subsection{Qualitative Results}\label{sec:qua}

\textbf{Module ablation}.  We present qualitative comparisons to illustrate the impact of various network configurations, as depicted in Fig.~\ref{fig:qualitative}. In the upper panel, contrasting C+UAO against C, we observe improvements due to the UAO's enhanced primitive understanding. For instance, an image initially misclassified as Bright-Orange under C correctly identifies as Draped-Silk with C+UAO. 
Further analysis between C+UAO and C+AO shows the benefits of incorporating object cues from top-$k$ predictions. This method rectifies attribute errors, such as correcting Ancient-Computer as Old-Computer, enhancing contextual relevance.
In comparisons of ULAO and C+UAO, we validate the effectiveness of the LAO's contrastive loss strategy. ULAO not only boosts the likelihood of accurate labeling but also effectively distinguishes between closely related labels and corrects mislinkages. For example, a table covered with draped fabric incorrectly labeled as Draped-Table by C+UAO is rightly adjusted to Draped-Fabric by ULAO, correcting the attribute-object linkage. Additionally, in the UT-Zappos dataset, ULAO improves the differentiating ability between similar materials like Sheepskin and Suede, underscoring the contrastive loss's role in handling subtle distinctions.

\textbf{Limitations and potentials}. The lower panel of Fig.~\ref{fig:qualitative} illustrates failure cases. Transitioning to UAO can lead to errors if the object cues are incorrect. 
Transitioning to ULAO, the model occasionally favors more specific or less typical labels over more general ones. For instance, ULAO classifies an image of Sliced-Fruit as Sliced-Apple. This tendency arises as general labels often receive high scores across many contexts, leading them to be frequently penalized by the adaptive threshold mechanism.  Some failure cases reveal some truth though: UAO interprets an image of an Ancient-Town as Straight-Street, recognizing the presence of a street within the town, and ULAO identifies Brushed-Fabric as resembling velvet. Such examples highlight that our model may have the potential to recognize multiple objects within a scene or refine generic labels into more precise ones.

\section{Conclusion}

This paper introduces a CLIP-based framework, Understanding and Linking Attributes and Objects (ULAO), tailored for Compositional Zero-Shot Learning. Our framework advances the state-of-the-art by strategically incorporating sequential object-to-attribute predictions, utilizing object predictions as contextual cues to enhance attribute recognition. Additionally, we innovate with a contrastive training strategy that generates hard negatives from the most confusing mispredictions. We refine this approach by implementing adaptive threshold adjustments that evolve as training progresses. This method not only improves the model’s ability to discriminate between similar primitives but also improves the accuracy of attribute-object linkages. Our evaluations across three datasets demonstrate ULAO's effectiveness, achieving superior performance in both Closed-World and Open-World scenarios.We also critically discuss model limitations and potential arising from the model design.

\bibliography{aaai25}

\begin{thebibliography}{38}
\providecommand{\natexlab}[1]{#1}

\bibitem[{Bao et~al.(2023)Bao, Chen, Huang, and Kong}]{bao2023prompting}
Bao, W.; Chen, L.; Huang, H.; and Kong, Y. 2023.
\newblock Prompting Language-Informed Distribution for Compositional Zero-Shot Learning.
\newblock \emph{arXiv preprint arXiv:2305.14428}.

\bibitem[{Doveh et~al.(2024)Doveh, Arbelle, Harary, Herzig, Kim, Cascante-Bonilla, Alfassy, Panda, Giryes, Feris et~al.}]{doveh2024dense}
Doveh, S.; Arbelle, A.; Harary, S.; Herzig, R.; Kim, D.; Cascante-Bonilla, P.; Alfassy, A.; Panda, R.; Giryes, R.; Feris, R.; et~al. 2024.
\newblock Dense and aligned captions (dac) promote compositional reasoning in vl models.
\newblock \emph{Advances in Neural Information Processing Systems}, 36.

\bibitem[{Hsieh et~al.(2023)Hsieh, Zhang, Ma, Kembhavi, and Krishna}]{hsieh2023sugarcrepe}
Hsieh, C.-Y.; Zhang, J.; Ma, Z.; Kembhavi, A.; and Krishna, R. 2023.
\newblock SugarCrepe: Fixing Hackable Benchmarks for Vision-Language Compositionality.
\newblock In \emph{Thirty-Seventh Conference on Neural Information Processing Systems Datasets and Benchmarks Track}.

\bibitem[{Huang et~al.(2024)Huang, Gong, Feng, Zhang, Lv, and Wang}]{Huang2024Troika}
Huang, S.; Gong, B.; Feng, Y.; Zhang, M.; Lv, Y.; and Wang, D. 2024.
\newblock Troika: Multi-Path Cross-Modal Traction for Compositional Zero-Shot Learning.
\newblock In \emph{Proceedings of the IEEE/CVF Conference on Computer Vision and Pattern Recognition}.

\bibitem[{Isola, Lim, and Adelson(2015)}]{isola2015discovering}
Isola, P.; Lim, J.~J.; and Adelson, E.~H. 2015.
\newblock Discovering states and transformations in image collections.
\newblock In \emph{Proceedings of the IEEE conference on computer vision and pattern recognition}, 1383--1391.

\bibitem[{Jiang and Zhang(2024)}]{jiang2024revealing}
Jiang, C.; and Zhang, H. 2024.
\newblock Revealing the Proximate Long-Tail Distribution in Compositional Zero-Shot Learning.
\newblock In \emph{Proceedings of the AAAI Conference on Artificial Intelligence}, volume~38, 2498--2506.

\bibitem[{Karthik, Mancini, and Akata(2022)}]{karthik2022kg}
Karthik, S.; Mancini, M.; and Akata, Z. 2022.
\newblock KG-SP: Knowledge Guided Simple Primitives for Open World Compositional Zero-Shot Learning.
\newblock In \emph{Proceedings of the IEEE/CVF Conference on Computer Vision and Pattern Recognition}, 9336--9345.

\bibitem[{Kingma and Ba(2014)}]{kingma2014adam}
Kingma, D.~P.; and Ba, J. 2014.
\newblock Adam: A method for stochastic optimization.
\newblock \emph{arXiv preprint arXiv:1412.6980}.

\bibitem[{Lake(2014)}]{lake2014towards}
Lake, B.~M. 2014.
\newblock \emph{Towards more human-like concept learning in machines: Compositionality, causality, and learning-to-learn}.
\newblock Ph.D. thesis, Massachusetts Institute of Technology.

\bibitem[{Li et~al.(2022)Li, Yang, Wei, Deng, and Yang}]{li2022siamese}
Li, X.; Yang, X.; Wei, K.; Deng, C.; and Yang, M. 2022.
\newblock Siamese Contrastive Embedding Network for Compositional Zero-Shot Learning.
\newblock In \emph{Proceedings of the IEEE/CVF Conference on Computer Vision and Pattern Recognition}, 9326--9335.

\bibitem[{Li et~al.(2024)Li, Liu, Chen, and Yao}]{li2024context}
Li, Y.; Liu, Z.; Chen, H.; and Yao, L. 2024.
\newblock Context-based and Diversity-driven Specificity in Compositional Zero-Shot Learning.
\newblock In \emph{Proceedings of the IEEE/CVF Conference on Computer Vision and Pattern Recognition}.

\bibitem[{Li et~al.(2023)Li, Liu, Jha, and Yao}]{li2023distilled}
Li, Y.; Liu, Z.; Jha, S.; and Yao, L. 2023.
\newblock Distilled reverse attention network for open-world compositional zero-shot learning.
\newblock In \emph{Proceedings of the IEEE/CVF International Conference on Computer Vision}, 1782--1791.

\bibitem[{Li et~al.(2020)Li, Xu, Mao, and Lu}]{li2020symmetry}
Li, Y.-L.; Xu, Y.; Mao, X.; and Lu, C. 2020.
\newblock Symmetry and group in attribute-object compositions.
\newblock In \emph{Proceedings of the IEEE/CVF Conference on Computer Vision and Pattern Recognition}, 11316--11325.

\bibitem[{Liu et~al.(2023)Liu, Li, Yao, Chang, Fang, Wu, and Saddik}]{liu2023pami}
Liu, Z.; Li, Y.; Yao, L.; Chang, X.; Fang, W.; Wu, X.; and Saddik, A.~E. 2023.
\newblock Simple Primitives with Feasibility- and Contextuality-Dependence for Open-World Compositional Zero-shot Learning.
\newblock \emph{IEEE Transactions on Pattern Analysis and Machine Intelligence}, 1--18.

\bibitem[{Lu et~al.(2023)Lu, Guo, Liu, and Guo}]{lu2023decomposed}
Lu, X.; Guo, S.; Liu, Z.; and Guo, J. 2023.
\newblock Decomposed soft prompt guided fusion enhancing for compositional zero-shot learning.
\newblock In \emph{Proceedings of the IEEE/CVF Conference on Computer Vision and Pattern Recognition}, 23560--23569.

\bibitem[{Mancini et~al.(2021)Mancini, Naeem, Xian, and Akata}]{mancini2021open}
Mancini, M.; Naeem, M.~F.; Xian, Y.; and Akata, Z. 2021.
\newblock Open world compositional zero-shot learning.
\newblock In \emph{Proceedings of the IEEE/CVF conference on computer vision and pattern recognition}, 5222--5230.

\bibitem[{Mancini et~al.(2022)Mancini, Naeem, Xian, and Akata}]{mancini2022learning}
Mancini, M.; Naeem, M.~F.; Xian, Y.; and Akata, Z. 2022.
\newblock Learning graph embeddings for open world compositional zero-shot learning.
\newblock \emph{IEEE Transactions on Pattern Analysis and Machine Intelligence}.

\bibitem[{Momeni et~al.(2023)Momeni, Caron, Nagrani, Zisserman, and Schmid}]{momeni2023verbs}
Momeni, L.; Caron, M.; Nagrani, A.; Zisserman, A.; and Schmid, C. 2023.
\newblock Verbs in action: Improving verb understanding in video-language models.
\newblock In \emph{Proceedings of the IEEE/CVF International Conference on Computer Vision}, 15579--15591.

\bibitem[{Naeem et~al.(2021)Naeem, Xian, Tombari, and Akata}]{naeem2021learning}
Naeem, M.~F.; Xian, Y.; Tombari, F.; and Akata, Z. 2021.
\newblock Learning graph embeddings for compositional zero-shot learning.
\newblock In \emph{Proceedings of the IEEE/CVF Conference on Computer Vision and Pattern Recognition}, 953--962.

\bibitem[{Nayak, Yu, and Bach(2023)}]{csp2023}
Nayak, N.~V.; Yu, P.; and Bach, S.~H. 2023.
\newblock Learning to Compose Soft Prompts for Compositional Zero-Shot Learning.
\newblock In \emph{International Conference on Learning Representations}.

\bibitem[{Purushwalkam et~al.(2019)Purushwalkam, Nickel, Gupta, and Ranzato}]{purushwalkam2019task}
Purushwalkam, S.; Nickel, M.; Gupta, A.; and Ranzato, M. 2019.
\newblock Task-driven modular networks for zero-shot compositional learning.
\newblock In \emph{Proceedings of the IEEE/CVF International Conference on Computer Vision}, 3593--3602.

\bibitem[{Radford et~al.(2021)Radford, Kim, Hallacy, Ramesh, Goh, Agarwal, Sastry, Askell, Mishkin, Clark et~al.}]{radford2021learning}
Radford, A.; Kim, J.~W.; Hallacy, C.; Ramesh, A.; Goh, G.; Agarwal, S.; Sastry, G.; Askell, A.; Mishkin, P.; Clark, J.; et~al. 2021.
\newblock Learning transferable visual models from natural language supervision.
\newblock In \emph{International conference on machine learning}, 8748--8763. PMLR.

\bibitem[{Saini, Pham, and Shrivastava(2022)}]{saini2022disentangling}
Saini, N.; Pham, K.; and Shrivastava, A. 2022.
\newblock Disentangling Visual Embeddings for Attributes and Objects.
\newblock In \emph{Proceedings of the IEEE/CVF Conference on Computer Vision and Pattern Recognition}, 13658--13667.

\bibitem[{Singh et~al.(2023)Singh, Zhang, Wang, Wang, Xiong, Du, and Chen}]{singh2023coarse}
Singh, H.; Zhang, P.; Wang, Q.; Wang, M.; Xiong, W.; Du, J.; and Chen, Y. 2023.
\newblock Coarse-to-Fine Contrastive Learning in Image-Text-Graph Space for Improved Vision-Language Compositionality.
\newblock \emph{arXiv preprint arXiv:2305.13812}.

\bibitem[{Thrush et~al.(2022)Thrush, Jiang, Bartolo, Singh, Williams, Kiela, and Ross}]{thrush2022winoground}
Thrush, T.; Jiang, R.; Bartolo, M.; Singh, A.; Williams, A.; Kiela, D.; and Ross, C. 2022.
\newblock Winoground: Probing vision and language models for visio-linguistic compositionality.
\newblock In \emph{Proceedings of the IEEE/CVF Conference on Computer Vision and Pattern Recognition}, 5238--5248.

\bibitem[{Wang et~al.(2023{\natexlab{a}})Wang, Yang, Wei, and Deng}]{wang2023hierarchical}
Wang, H.; Yang, M.; Wei, K.; and Deng, C. 2023{\natexlab{a}}.
\newblock Hierarchical Prompt Learning for Compositional Zero-Shot Recognition.
\newblock In \emph{IJCAI}.

\bibitem[{Wang et~al.(2023{\natexlab{b}})Wang, Liu, Jing, Chen, Liang, Wang, and Shen}]{wang2023learning}
Wang, Q.; Liu, L.; Jing, C.; Chen, H.; Liang, G.; Wang, P.; and Shen, C. 2023{\natexlab{b}}.
\newblock Learning Conditional Attributes for Compositional Zero-Shot Learning.
\newblock In \emph{Proceedings of the IEEE/CVF Conference on Computer Vision and Pattern Recognition}, 11197--11206.

\bibitem[{Wei et~al.(2019)Wei, Yang, Wang, Deng, and Liu}]{wei2019adversarial}
Wei, K.; Yang, M.; Wang, H.; Deng, C.; and Liu, X. 2019.
\newblock Adversarial fine-grained composition learning for unseen attribute-object recognition.
\newblock In \emph{Proceedings of the IEEE/CVF International Conference on Computer Vision}, 3741--3749.

\bibitem[{Xu, Chai, and Kordjamshidi(2024)}]{xu2024gipcol}
Xu, G.; Chai, J.; and Kordjamshidi, P. 2024.
\newblock GIPCOL: Graph-Injected Soft Prompting for Compositional Zero-Shot Learning.
\newblock In \emph{Proceedings of the IEEE/CVF Winter Conference on Applications of Computer Vision}, 5774--5783.

\bibitem[{Xu, Kordjamshidi, and Chai(2021)}]{xu2021zero}
Xu, G.; Kordjamshidi, P.; and Chai, J.~Y. 2021.
\newblock Zero-shot compositional concept learning.
\newblock \emph{arXiv preprint arXiv:2107.05176}.

\bibitem[{Xu et~al.(2021)Xu, Wang, Wong, and Kankanhalli}]{xu2021relation}
Xu, Z.; Wang, G.; Wong, Y.; and Kankanhalli, M.~S. 2021.
\newblock Relation-aware Compositional Zero-shot Learning for Attribute-Object Pair Recognition.
\newblock \emph{IEEE Transactions on Multimedia}.

\bibitem[{Yang et~al.(2023)Yang, Pan, Li, Yang, and Deng}]{yang2023dual}
Yang, Y.; Pan, R.; Li, X.; Yang, X.; and Deng, C. 2023.
\newblock Dual-stream contrastive learning for compositional zero-shot recognition.
\newblock \emph{IEEE Transactions on Multimedia}.

\bibitem[{Yu and Grauman(2014)}]{yu2014fine}
Yu, A.; and Grauman, K. 2014.
\newblock Fine-grained visual comparisons with local learning.
\newblock In \emph{Proceedings of the IEEE conference on computer vision and pattern recognition}, 192--199.

\bibitem[{Yu and Grauman(2017)}]{yu2017semantic}
Yu, A.; and Grauman, K. 2017.
\newblock Semantic jitter: Dense supervision for visual comparisons via synthetic images.
\newblock In \emph{Proceedings of the IEEE International Conference on Computer Vision}, 5570--5579.

\bibitem[{Yuksekgonul et~al.(2022)Yuksekgonul, Bianchi, Kalluri, Jurafsky, and Zou}]{yuksekgonul2022and}
Yuksekgonul, M.; Bianchi, F.; Kalluri, P.; Jurafsky, D.; and Zou, J. 2022.
\newblock When and why vision-language models behave like bags-of-words, and what to do about it?
\newblock In \emph{The Eleventh International Conference on Learning Representations}.

\bibitem[{Zhang, Awal, and Agrawal(2024)}]{zhang2024contrasting}
Zhang, L.; Awal, R.; and Agrawal, A. 2024.
\newblock Contrasting Intra-Modal and Ranking Cross-Modal Hard Negatives to Enhance Visio-Linguistic Compositional Understanding.
\newblock In \emph{Proceedings of the IEEE/CVF Conference on Computer Vision and Pattern Recognition}, 13774--13784.

\bibitem[{Zheng, Zhu, and Nevatia(2024)}]{zheng2024caila}
Zheng, Z.; Zhu, H.; and Nevatia, R. 2024.
\newblock CAILA: Concept-Aware Intra-Layer Adapters for Compositional Zero-Shot Learning.
\newblock In \emph{Proceedings of the IEEE/CVF Winter Conference on Applications of Computer Vision}, 1721--1731.

\bibitem[{Zhou et~al.(2022)Zhou, Yang, Loy, and Liu}]{zhou2022learning}
Zhou, K.; Yang, J.; Loy, C.~C.; and Liu, Z. 2022.
\newblock Learning to prompt for vision-language models.
\newblock \emph{International Journal of Computer Vision}, 130(9): 2337--2348.

\end{thebibliography}

\end{document}